\relax
\documentclass[letterpaper]{article}

\usepackage{times}
\usepackage{helvet}
\usepackage{courier}
\usepackage[hyphens]{url}
\usepackage{graphicx}
\usepackage{natbib}
\usepackage{caption}
\DeclareCaptionStyle{ruled}{labelfont=normalfont,labelsep=colon,strut=off}
\usepackage{cite}
\usepackage{amsmath,amssymb,amsfonts}
\usepackage{multirow,adjustbox}
\usepackage{lipsum}
\usepackage{soul,color} 

\title{Spiking Neural Networks with Improved Inherent Recurrence Dynamics for Sequential Learning}
\author {
    Wachirawit Ponghiran, Kaushik Roy\\
}

\begin{document}

\maketitle

\begin{abstract}
Spiking neural networks (SNNs) with leaky integrate and fire (LIF) neurons, can be operated in an event-driven manner and have internal states to retain information over time, providing opportunities for energy-efficient neuromorphic computing, especially on edge devices.
Note, however, many representative works on SNNs do not fully demonstrate the usefulness of their inherent recurrence (membrane potential retaining information about the past) for sequential learning. Most of the works train SNNs to recognize static images by artificially expanded input representation in time through rate coding. 
We show that SNNs can be trained for sequential tasks and propose modifications to a network of LIF neurons that enable internal states to learn long sequences and make their inherent recurrence resilient to the vanishing gradient problem.
We then develop a training scheme to train the proposed SNNs with improved inherent recurrence dynamics. Our training scheme allows spiking neurons to produce multi-bit outputs (as opposed to binary spikes) which help mitigate the mismatch between a derivative of spiking neurons' activation function and a surrogate derivative used to overcome spiking neurons' non-differentiability.
Our experimental results indicate that the proposed SNN architecture on TIMIT and LibriSpeech 100h dataset yields accuracy comparable to that of LSTMs (within $1.10\%$ and $0.36\%$, respectively), but with $2\times$ fewer parameters than LSTMs. The sparse SNN outputs also lead to $10.13\times$ and $11.14\times$ savings in multiplication operations compared to GRUs, which is generally considered as a lightweight alternative to LSTMs, on TIMIT and LibriSpeech 100h datasets, respectively.
\end{abstract}


\section{Introduction}

Spiking neural networks (SNNs) have been considered as a promising solution for low-power neuromorphic computing. Inspired by the biological neural networks, neurons in the SNNs communicate asynchronously with each other through binary values that aim to represent the actual neuron spiking activity. On event-driven hardware, the asynchronous binary communication of the neurons leads to potential power savings by requiring computation units to be activated only when the spikes occur. This makes SNNs appealing for applications on edge devices. As operations on mobile/edge devices are typically limited by power/energy constraints, 
exploiting common techniques to improve efficiency such as batch computation may not always be possible. SNNs provide an opportunity to process inputs sequentially at low-power consumption by computing in an event-driven manner. In addition, SNNs also have internal states (membrane potentials) to retain information over time. Such inherent recurrence in SNNs is suitable for sequential learning, but its usefulness still has not been well demonstrated on suitable applications.

\begin{figure}[!t]
\begin{center}
  \includegraphics[width=\linewidth]{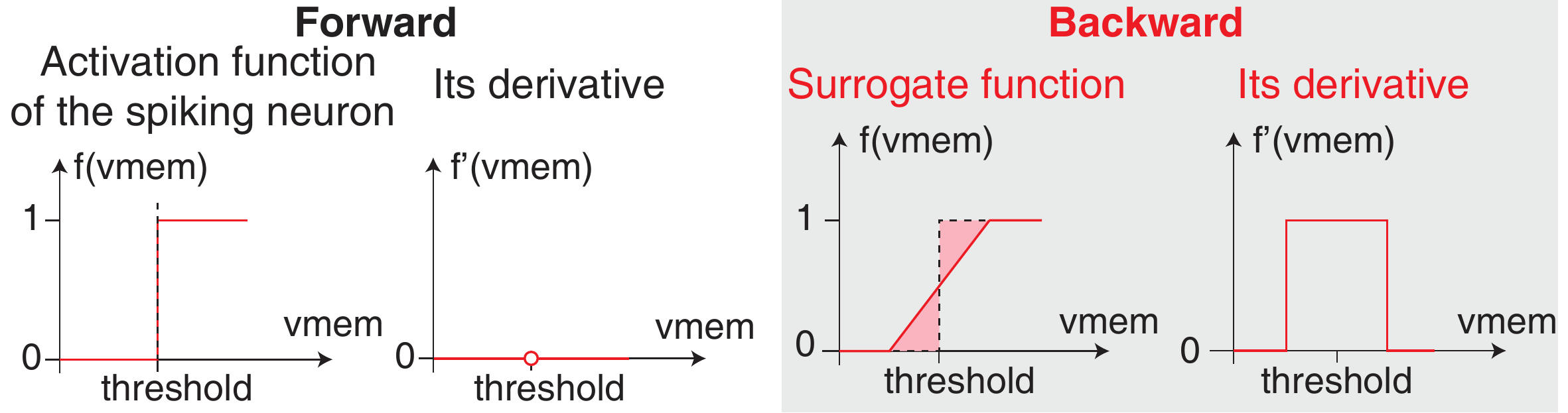}
\end{center}
\caption{Technique to overcome non-differentiability of the spiking neurons with a straight-through estimator. The activation function of the spiking neuron with an ill-defined derivative (left) is replaced with a surrogate function (right) during backward pass. The substitution enables the use of backpropagation to train SNNs; however, it leads to a discrepancy between actual and surrogate gradients. The gradient mismatch results in noisy gradient updates and consequently degrades learning performance. The shaded area highlights the difference between the activation function of the spiking neuron and its surrogate function. The role of threshold on generating outputs is discussed in Section~\ref{sec:background}.
}
\label{fig:intro_gradmismatch_1}
\end{figure}

Many prior works on training SNNs for energy-efficient computing focus on recognition problems from static images, where the inputs are expanded in time through rate coding~\citep{diehl2015fast,rueckauer2017conversion,sengupta2019going}. Each pixel is mapped to a binary sequence where the occurrence of ones in the sequence is proportional to the pixel intensity. Even though the rate-coded inputs are sequential, such synthetic tasks neither signify the importance of learning  "time" with SNN's inherent recurrence, nor represent practical sequential learning problems. Recently, \citep{cramer2020heidelberg} has successfully trained a network of leaky integrate-and-fire (LIF) spiking neurons with back-propagation through time (BPTT) for a sequential recognition task. Nonetheless, the task is small-vocabulary keyword spotting for spoken digits. Training SNNs for larger sequential tasks remains a difficult problem. We show that the issue partially stems from vanishing gradients. Gradients that propagate along synaptic currents and membrane potentials, which are both neuron internal states, quickly diminish with time-steps. Note each time-step represents how far in the input sequences that have been processed by the SNN. Since we do not expand each input sequence in time, a total number of time-steps is equivalent to the length of input sequences.

In this work, we propose SNNs with improved inherent recurrence dynamics to effectively learn long sequences. We identify that gradients mostly flow along the synaptic currents and a constant decay of the synaptic currents is a cause of diminishing gradients. Simply making the synaptic currents retain values without any leak does not eliminate the problem and in fact, leads to uncontrollable growth in the values during the forward pass.
Hence, we introduce the concept of "forget gate" as in the gate-recurrent units (GRUs) and long-short term memory units (LSTMs) to enable selective updates of the synaptic currents during forward pass. This is an empirically shown method to sustain gradients through multiple time-steps~\citep{hochreiter1997long,cho2014properties}. We show that the proposed modifications lead to improvement in speech recognition accuracy on both TIMIT and LibriSpeech 100h dataset. 
We also come up with a scheme to train the proposed SNNs and at the same time address the gradient mismatch problem which arises from the use of surrogate gradient during training. The trick to overcome the non-differentiability of spiking neurons with a straight-through estimator~\citep{courbariaux2016binarized} leads to a discrepancy between the presumed and the actual activation function as shown in Fig.~\ref{fig:intro_gradmismatch_1}. This mismatch results in noisy gradient updates and has been shown to affect the learning performance~\citep{lin2016overcoming}. To alleviate this problem, we extend spiking neurons to output multi-bit outputs (as opposed to binary spikes). We follow observations from the existing literature that increasing output precision helps reduce gradient mismatch~\citep{lin2016overcoming,kim2020binaryduo}. We show that the learning performance of SNNs with multi-bit outputs is better than the network with binary outputs. 
Importantly, we demonstrate that the proposed SNNs achieve similar speech recognition accuracy to the LSTMs on TIMIT and LibriSpeech 100h dataset but with only half the number of parameters.
The benefit of SNNs that communicate sparsely with many zero messages leads to savings in the number of multiplication operations, which govern the overall operations.

The remainder of the paper is organized as follows. Section~\ref{sec:background} provides background on the spiking LIF neurons and a method to train them using BPTT. We then discuss the potential benefit of executing them on an event-driven platform. Section~\ref{sec:method} shows the vanishing gradient problem that arises when a layer of LIF neurons is used to learn long sequences. We then propose SNNs with improved inherent recurrence dynamics for mitigating the diminishing gradients and present a method to train them. Experimental results are presented in Section~\ref{sec:experiment}, where we show the effectiveness of the proposed network in learning and compare accuracy and computation savings to LSTMs and GRUs. The paper is concluded in Section~\ref{sec:conclusion}.


\section{Background} \label{sec:background}

In this section, we introduce the dynamics of LIF neurons, a popular spiking neuron model in the field of computational neuroscience, and explain a method to train a network of the LIF neurons using BPTT.
We then discuss the benefit of executing SNNs (that produce many zero outputs) on suitable event-driven hardware.

\begin{figure}[!t]
\begin{center}
\includegraphics[width=\linewidth]{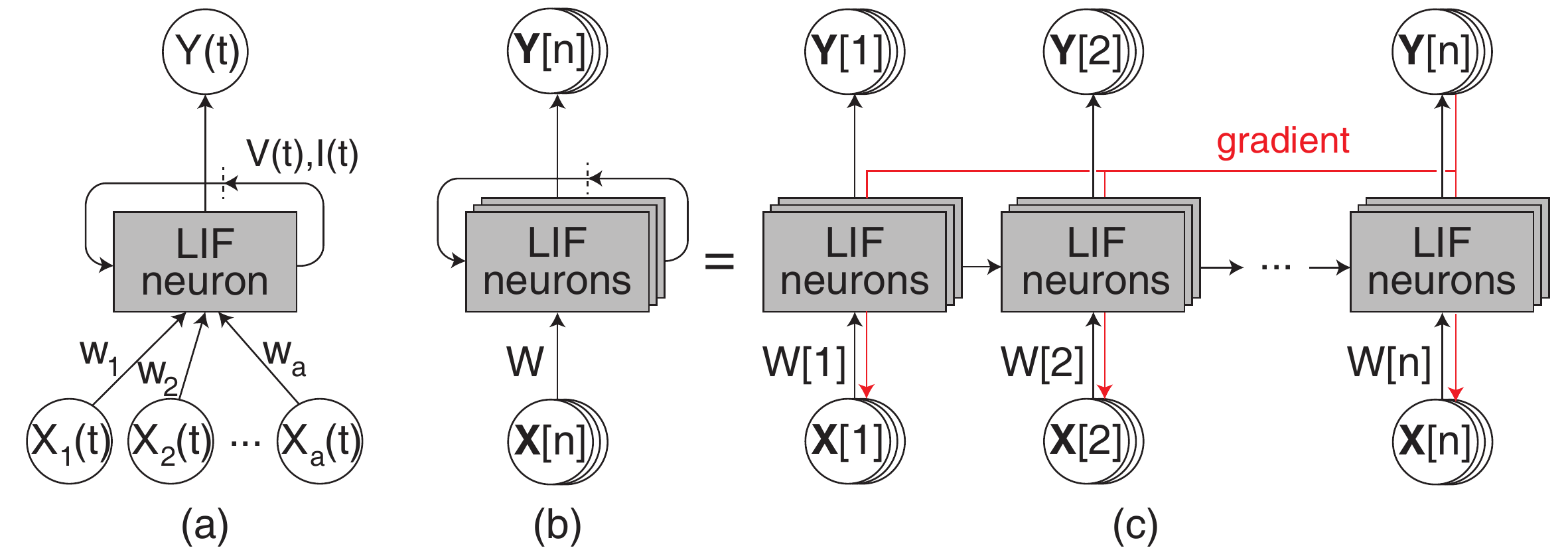}
\end{center}
\caption{(a) An example of LIF neuron which receives $X_i(t)$ as inputs and produces $Y(t)$ as an output. $w_i$ denotes the weight of each connection to the neuron. (b) Diagram representing operations of spiking neurons that form a neuron layer in a discrete-time. The neurons receive input vector $\mathbb{X}[n]$ and produce output vector $\mathbb{Y}[n]$ at time-step $n$. (c) Equivalent representation when the operations are unrolled into $n$ time-steps.}
\label{fig:background_neurondyn}
\end{figure}

\subsection{Dynamics of LIF Neurons}

Suppose an LIF neuron receives input $X_i(t)$ from input neuron $i{\in}{ \{1,2,...,a\}}$ at time $t$ as shown in Fig.~\ref{fig:background_neurondyn}(a). Each input would go through a synapse and gets modulated with the associated weight $w_i$ before being integrated into a synaptic current $I(t)$ that decays exponentially according to a constant ${\tau_{syn}}$. Note, $I(t)$ is one internal state of the spiking neuron and behaves like a leaky integrator of the inputs. We can express it mathematically as:
\begin{align}
    \frac{dI(t)}{dt} = -\frac{I(t)}{\tau_{syn}} + \sum_{i=1}^{a}{w_{i}X_{i}(t)} \label{eq:lif_syn}
\end{align}
The increase in synaptic current leads to an increase in membrane potential $V(t)$, another internal state of the spiking neuron. If the membrane potential crosses a firing threshold $\gamma$, the neuron produces an output spike. Otherwise, the membrane potential decays exponentially with a time constant ${\tau_{mem}}$. We can express the dynamics of the neuron as:
\begin{align}
    \frac{dV(t)}{dt} = -\frac{V(t)}{\tau_{mem}} + \eta I(t) - \gamma Y(t) \label{eq:lif_mem}
\end{align}
where $\eta$ represents an input resistance and $Y(t)$ is an output of the neuron at time $t$. Since the output of the spiking neuron is constrained to a binary value, $Y(t) = 1$ indicates that the neuron produces an output spike at time $t$. The last term of the Eq.~\ref{eq:lif_mem} accounts for an instantaneous decrease in the membrane potential after the neuron produces output. For simplicity, we assume input resistance ($\eta$) is 1 throughout this work.

Computing neuron states by directly solving the above equations is expensive, and hence, it is generally done by discretely evaluating the equations over small time-steps. The neuronal dynamics can be then approximated as:
\begin{align}
    I[n] &= \beta I[n-1] + \sum_{i=1}^{a}{w_{i}X_{i}[n]}  \label{eq:dlif_syn} \\
    V[n] &= \alpha V[n-1] + I[n] - \gamma Y[n-1] \label{eq:dlif_mem} \\
    Y[n] &= \mathrm{U}( V[n] - \gamma) \label{eq:dlif_out}
\end{align}
where $n$ represents an index of simulation time-step. $\beta$ and $\alpha$ are decaying factors for the synaptic current and the membrane potential, respectively. $\mathrm{U}(\cdot)$ denotes a unit step function:
\begin{align}
    \mathrm{U}(x) &= 
        \begin{cases}
        1 & \text{if}~x > 0 \\
        0 & \text{if}~x \leqslant 0
        \end{cases}  \label{eq:unit_step}
\end{align}

From Eq.~\ref{eq:dlif_out}-\ref{eq:unit_step}, threshold is a hyperparameter that dictates a distribution of spiking neuron outputs. Setting threshold to zero leads to a neuron that always spikes (i.e. always generates one as output), thus making the neuron not useful for learning. Having high threshold value leads to many zero outputs.
However, excessive zero outputs may harm learning as only connections to active neuron get updated in the backward pass. Hence, threshold has to be chosen carefully. We discuss a method for threshold selection that simplifies the SNN training in the next section.

\subsection{Training a Network of LIF Neurons with BPTT}
To understand the training process, let's continue with the previous example. We denote a vector of inputs as $\mathbb{X}[n]=(X_1[n], X_2[n], ...,  X_b[n])$. Suppose there are other $b-1$ spiking neurons that also receive input vector $\mathbb{X}[n]$
and together form a neuron layer. We represent operations of this neuron layer at time-step $n$ as illustrated in the Fig.~\ref{fig:background_neurondyn}(b). We denote a vector of outputs as $\mathbb{Y}[n]=(Y_1[n], Y_2[n], ...,  Y_b[n])$ given that $Y_j[n]$ is an output of neuron $j$ at time-step $n$. We introduce a connection matrix $\mathbb{W}$ which has an element $\mathbb{W}_{ij}$ in row $i$ and column $j$ indicating a weight of the connection between neuron $i$ and $j$. The dynamics of the neuron layer in a discrete time can then be expressed as:
\begin{align}
    \mathbb{I}[n] &= \beta \mathbb{I}[n-1] + {\mathbb{X}[n]}\mathbb{W} \label{eq:dlif_syn_vec} \\
    \mathbb{V}[n] &= \alpha \mathbb{V}[n-1] + \mathbb{I}[n] - \gamma \mathbb{Y}[n-1] \label{eq:dlif_mem_vec} \\
    \mathbb{Y}[n] &= \mathrm{U}(\mathbb{V}[n] - \gamma) \label{eq:dlif_out_vec}
\end{align}
Since the internal states of the neurons (i.e. $\mathbb{V}[n]$ and $\mathbb{I}[n]$) depend on the current inputs and their previous values at the last time-step, we can recursively apply the equations and BPTT algorithm to compute gradients. The operations of the neuron layer are unfolded in time by creating several copies of the LIF neurons and treat them as a feed-forward network with tied weights such that $\mathbb{W}[n]$ is treated as $\mathbb{W}$ for every time-step $n$. Fig.~\ref{fig:background_neurondyn}(c) shows the operations after the unrolling. The gradient can be computed and propagated backward through each time-step after completing the forward pass.
During the backward pass, the non-differentiable behavior of the unit step function is overcome by substituting the ill-defined differential with a smooth function~\citep{zenke2018superspike,bellec2018long,shrestha2018slayer}. Such concept is commonly known as straight-through estimator. The use of surrogate function is popular as it does not require any change to the optimization algorithm. 

\subsection{Benefits of Computing with SNNs}

Benefits of computing with SNNs comes in two different ways. One is the reduction in the number of parameters and another is a computational saving resulting from sparse inputs which ideally incur little or no energy consumption on an event-driven hardware. To quantify the benefits, we compare SNNs with GRUs and LSTMs, which are recurrent neural networks widely used for sequential learning.

Let us first consider parameter reduction. Suppose an input and output vector have a size of $m$ and $n$, respectively. The number of parameters for SNNs is $(m \cdot n)$ (previous subsection). GRUs and LSTMs have higher number of parameters as they have multiple signals to control the flow of information through them (see the appendix for more details). Specifically, GRUs have $3 (m \cdot n+n^2)$ parameters while LSTMs have $4 (m \cdot n+n^2)$ parameters. The difference in the number of parameters leads to reduction in memory. 

To demonstrate computational saving, we compute a cost for the neuron layer to update its internal states and generate outputs given a sparse input vector. We assume that the input vector to spiking neuron is sparse because we stack the spiking neuron layer to form a multiple-layer network and increase network capacity. Outputs from each layer are typically full of zeros as a result of the threshold function as in Eq.~\ref{eq:dlif_out_vec}, and the outputs from one layer are fed to the next layer as inputs. We then compare it against computation in GRUs and LSTMs given a dense input vector. Let $C_{sparse}$ denotes the cost of sparse computing with SNNs. Eq.~\ref{eq:dlif_syn_vec}-\ref{eq:dlif_out_vec} involve fetching synaptic weights, matrix-vector multiplications, vector-constant multiplications, and vector-vector additions. Actual cost for each operation depends strongly on the underlying hardware and implementation. For algorithmic development purpose, we assume that a suitable implementation minimize the memory access and only focus on arithmetic operations. We use number of multiplications which mainly govern overall operations as a cost metric and have:
\begin{align}
    C_{sparse} = \underbrace{(n + s \cdot m \cdot n)}_{\text{\# of multiplications for updating }\mathbb{I}[n]\text{,}}  + \underbrace{(3n)}_{\text{updating }\mathbb{V}[n]}
\end{align}
where $s$ represents a ratio of zeros in the input vector. Following the same logic, the cost of computing with GRUs and LSTMs are $3n + 3(m +n)n$ and $3n + 4(m +n)n$. Hence, if inputs and outputs have the same dimension, the computation saving is roughly be $6/s$ and $8/s$ for GRUs and LSTMs. 
This significant computational reduction of SNNs potentially leads to faster execution and less energy consumption.


\section{Vanishing Gradient Problem and\\the Proposed SNNs with Improved\\Inherent Recurrence Dynamics} \label{sec:method}

\subsection{Vanishing Gradient Problem}

\begin{figure}[!t]
\begin{center}
\includegraphics[width=\linewidth]{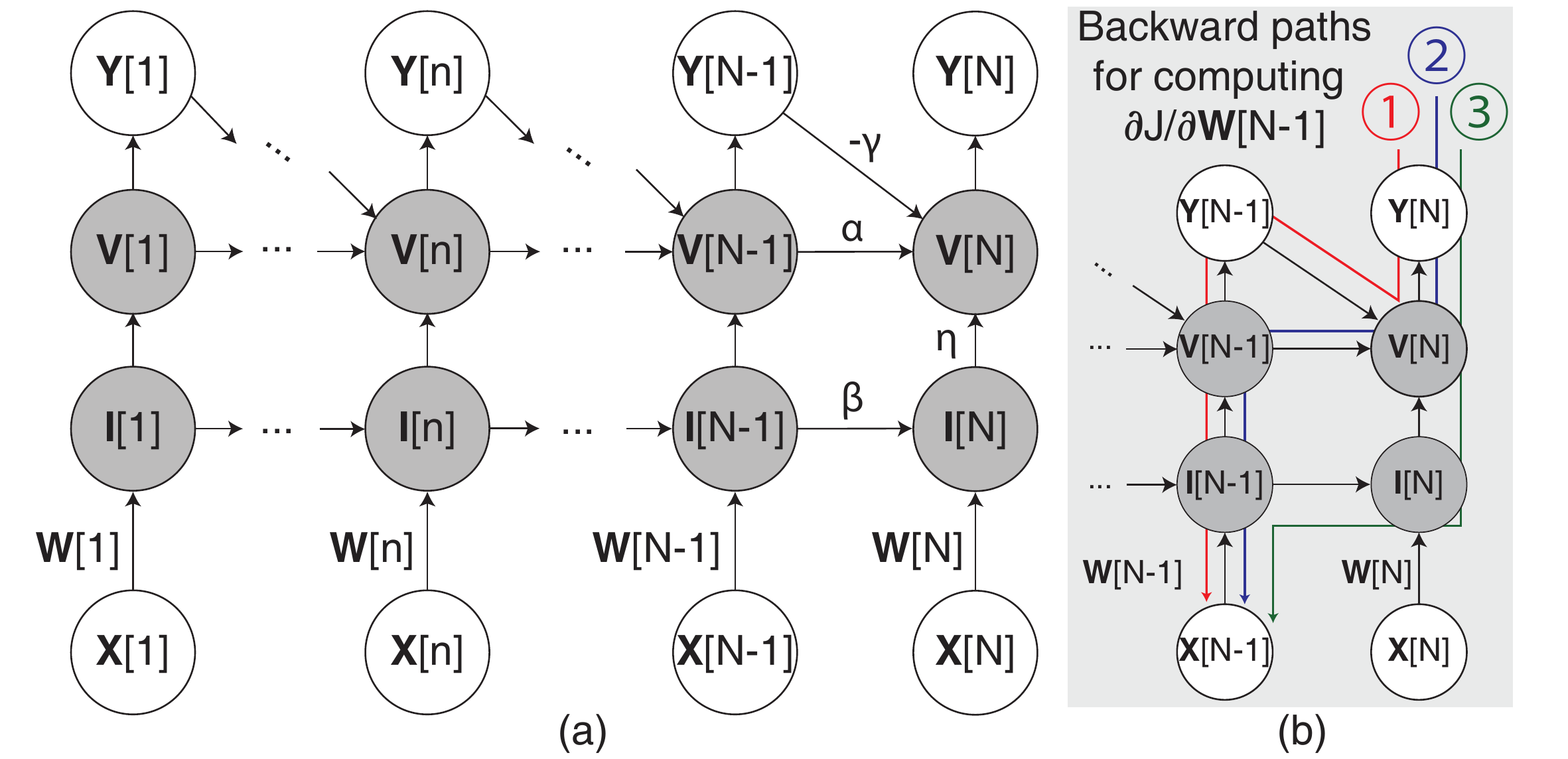}
\end{center}
\caption{(a) Computational graph of the spiking neuron layer. Input vector $\mathbb{X}[n]$ is sequentially fed to the neuron layer for $N$ time-steps.
Synaptic currents $\mathbb{I}[n]$ are updated based on the input vector, and the new values are added to membrane potentials $\mathbb{V}[n]$.
Output vector $\mathbb{Y}[n]$ is then generated according to Eq.~\ref{eq:dlif_out_vec}. Finally, values of synaptic currents and membrane potentials are carried over to the next time-step while decaying with a factor $\beta$ and $\alpha$. Feedback connection from outputs to membrane potentials lead to a reduction in the membrane potentials whenever non-zero outputs are generated. (b) Graph demonstrating all backward paths associated with computing gradient $\partial J/\partial \mathbb{W}[N-1]$.}
\label{fig:methods_compgraph}
\end{figure}

To demonstrate the vanishing gradient problem of the spiking neuron layer, we derive a gradient for updating connection matrix $\mathbb{W}$ based on the computational graph illustrated in Fig.~\ref{fig:methods_compgraph}(a).
Suppose $J$ is an objective that we would like to minimize at time-step $N$. The equation for calculating the gradient is:
\begin{align}
    \frac{\partial J}{\partial \mathbb{W}} &= \sum_{n=1}^{N} \frac{\partial J}{\partial \mathbb{W}[n]} = \sum_{n=1}^{N} \frac{\partial J}{\partial \mathbb{V}[N]} \frac{\partial \mathbb{V}[N ]}{\partial \mathbb{W}[n]}  \label{eq:lif_bptt}
\end{align}

Since we treat all $\mathbb{W}[n]$ as the same weight matrix, 
gradient for updating $\mathbb{W}$ is a summation of all ${\partial J}/{\partial \mathbb{W}[n]}$ from all time-steps. To make sure that the gradient does not vanish, the term $\mathbb{V}[N]/\mathbb{W}[n]$ should remain significant. However, we show that the term diminishes for large value of $N-n$. Plugging Eq.~\ref{eq:dlif_mem_vec} and~\ref{eq:dlif_out_vec} into the last term of Eq.~\ref{eq:lif_bptt} gives:

\begin{align}
    \frac{\partial \mathbb{V}[N]}{\partial \mathbb{W}[n]} &= \frac{\partial \left(\alpha \mathbb{V}[N-1] + \mathbb{I}[N] - \gamma \mathrm{U} ( \mathbb{V}[N-1]  - \gamma) \right)}{\partial \mathbb{W}[n]} \nonumber \\
    &= P[N-1] \frac{\partial \mathbb{V}[N-1]}{\partial \mathbb{W}[n]} + \frac{\partial \mathbb{I}[N]}{\partial \mathbb{W}[n]}
\end{align}
where
\begin{align}
    P[n] &= \frac{\partial \left(\alpha \mathbb{V}[n] - \gamma \mathrm{U} ( \mathbb{V}[n]  - \gamma)\right)}{\partial\mathbb{V}[n]} \label{eq:term_p}
\end{align}
Continuing substitution $\mathbb{V}[n]$ leads to a closed-form expression:
\begin{align}
    \frac{\partial \mathbb{V}[N]}{\partial \mathbb{W}[n]} &= \left( \prod_{k=1}^{N-n} P[N-k] \right) \frac{\partial \mathbb{V}[n]}{\partial \mathbb{W}[n]} \nonumber \\
    & \quad + \sum_{l=1}^{N-(n+1)} \left( \prod_{k=1}^{l} P[N-k] \right)  \frac{\partial \mathbb{I}[N-l]}{\partial \mathbb{W}[n]} \nonumber \\
    & \quad + \frac{\partial \mathbb{I}[N]}{\partial \mathbb{W}[n]} \label{eq:lif_closeform}
\end{align}

Intuitively, the expression above is equivalent to summing partial gradients through all possible backward paths from $\mathbb{V}[N]$ to $\mathbb{W}[n]$. Fig.~\ref{fig:methods_compgraph}(b) shows an example when $n=N-1$. There are 3 possible backward paths from $\mathbb{V}[N]$ to $\mathbb{W}[N-1]$, i.e. \textcircled{1} through $\mathbb{Y}[N-1]$, \textcircled{2} directly through $\mathbb{V}[N-1]$, and \textcircled{3} through $\mathbb{I}[N]$. With a piecewise linear surrogate function, $P[n]$ becomes a term with a small value. Hence, the last term in Eq.~\ref{eq:lif_closeform} remains the most significant term implying that the gradients mostly flow along the synaptic currents. Substituting Eq.~\ref{eq:dlif_syn_vec} into the last term gives:
\begin{align}
    \frac{\partial \mathbb{I}[N]}{\partial \mathbb{W}[n]} &= \frac{\partial \mathbb{I}[N]}{\partial \mathbb{I}[N-1]} \frac{\partial \mathbb{I}[N-1]}{\partial \mathbb{I}[N-2]} \cdots
    \frac{\partial \mathbb{I}[n]}{\partial \mathbb{W}[n]} \nonumber \\
    & = \left( \prod_{k=n}^{N-1} \frac{\partial \mathbb{I}[k+1]}{\partial \mathbb{I}[k]} \right) \mathbb{X}[n] = \beta^{N-n} \mathbb{X}[n] \label{eq:deriv_syn}
\end{align}
As decaying rate $\beta$ is typically a constant and set to a value less than $1$, Eq.~\ref{eq:deriv_syn} converges to zero for sufficiently large $N-n$. This indicates that the spiking neuron layer suffers from a vanishing gradient problem which prevents it from learning a long sequence. Note, however, we can modify the internal state update equations of the neuron layer to avoid the problem. The following subsection talks about internal states of the spiking neuron layer and the proposed modifications in details.

\subsection{Improving Inherent Recurrence Dynamics}

There are two distinct internal states of the spiking neuron layer as discussed earlier: synaptic currents and membrane potentials.
Synaptic currents carry information over time as they are the main path for gradients to flow backward. Membrane potentials keep track of errors when the spiking neurons produce outputs. Since spiking neurons have a threshold, outputs are not always transmitted, leading to their event-driven characteristic. Keeping track of the blocked outputs (so-called residual errors) was shown to improve performance when stacking the spiking neurons into multiple layers~\citep{rueckauer2017conversion}. To retain significant gradients for a long time which is one of the focus in this study, we only modify the synaptic current equation while leaving the equation for updating membrane potential the same.

From Eq.~\ref{eq:dlif_syn_vec}, note that the synaptic currents of the spiking neurons are multiplied by the same constant every time-step. This leads to fast diminishing gradients during backward pass as we have shown in the previous subsection.
Selectively updating neuron states like GRUs and LSTMs is an empirical method to maintain magnitudes of the gradients. We follow the same technique to update synaptic currents and propose equations for updating synaptic currents as follows:
\begin{align}
    \mathbb{F}[n] &= \sigma ({\mathbb{X}[n]}\mathbb{W}_{fi}) \label{eq:forget_1} \\
    \mathbb{C}[n] &= ReLU({\mathbb{X}[n]}\mathbb{W}_{ci}) \label{eq:candidate_1} \\
    \mathbb{I}[n] &= \mathbb{F}[n] \odot \mathbb{I}[n-1] + (1-\mathbb{F}[n]) \odot \mathbb{C}[n] \label{eq:new_syn_eq}
\end{align}
where $\odot$ signifies element-wise multiplication. $\mathbb{W}_{fi}$ and $\mathbb{W}_{ci}$ are weight matrices for input messages to determine forget and candidate signal, respectively. As in GRUs and LSTMs, the forget signal controls amount of information passed from the previous time-step. The candidate signal represents potential information to be added to the synaptic currents at each time-step. 
Throughout this work, we call SNNs after this modification as ``the SNNs with improved dynamics v1" for short.

From the Eq.~\ref{eq:new_syn_eq}, updates on the synaptic currents are determined solely based on the inputs at every time-step regardless of the existing values.
The existing values of the synaptic currents may provide useful information about the past and overwriting them can be avoided by using previous synaptic current for forget and candidate signal computation. In other words, we may choose to compute forget signals based on the present inputs and the previous synaptic currents (denoted by $\mathbb{I}[n-1]$). However, doing so would eliminate the benefit of having sparse inputs. This is because $\mathbb{I}[n-1]$ is a dense vector. We can use $\mathbb{Y}[n]$ instead as it is sparse and provides an approximation of $\mathbb{I}[n]$. Similar logic also applies to the computation of the candidate signal. Therefore, we propose a new set of equations to update synaptic currents as follows:
\begin{align}
    \mathbb{F}[n] &= \sigma ({\mathbb{X}[n]}\mathbb{W}_{fi} + {\mathbb{Y}[n-1]}\mathbb{W}_{fr}) \\
    \mathbb{C}[n] &= ReLU({\mathbb{X}[n]}\mathbb{W}_{ci} + {\mathbb{Y}[n-1]}\mathbb{W}_{cr})  \\
    \mathbb{I}[n] &= \mathbb{F}[n] \odot \mathbb{I}[n-1] + (1-\mathbb{F}[n]) \odot \mathbb{C}[n] 
\end{align}
where $\mathbb{W}_{fr}$ and $\mathbb{W}_{cr}$ are weight matrices for previous outputs to determine forget and candidate signal, respectively. We refer to SNNs after this modification as ``the SNNs with improved dynamics v2". We later show that the proposed modification further leads to higher recognition accuracy. The cost of computation using the proposed spiking neuron layer is increased roughly fourfold because the modifications lead to $4\times$ more matrix-vector multiplications than the original spiking neuron layer.
However, we show in the results section that proper selection of threshold in SNNs lead to sparse outputs, which ideally incur few computations on event-driven hardware. Thus, we can still expect benefit of sparse communications during inference. 

\subsection{Training Method for the Proposed SNNs with Improved Inherent Recurrence Dynamics}

\begin{figure}[!t]
\begin{center}
\includegraphics[width=\linewidth]{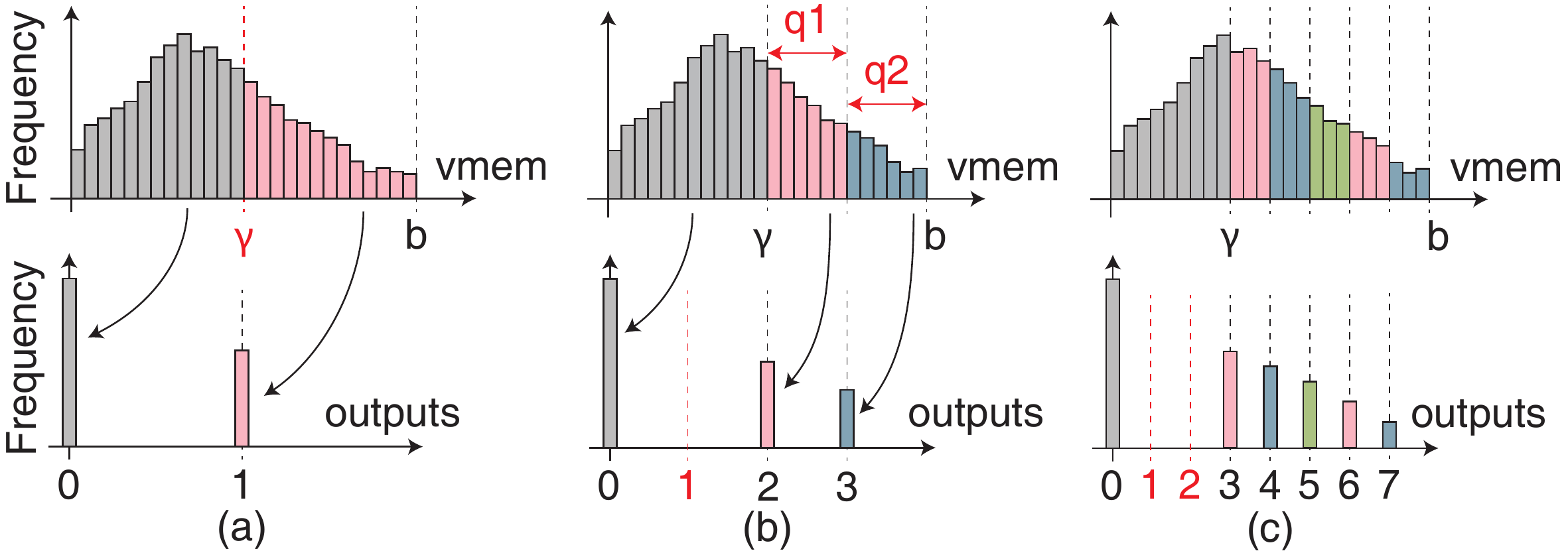}
\end{center}
\caption{(a) Activation function of the spiking neuron maps membrane potentials, which are real values, to binary outputs. However, the mapping depends on the choice of threshold ($\gamma$) and the maximum value of membrane potentials ($b$). Existing works generally set threshold as a constant prior to training. (b) The activation function after a modification to produce 2-bit outputs. Threshold is set to $2b/4$; values below the threshold are mapped to zeros. (c) The activation function after a modification to produce 3-bit outputs. Threshold is set to $3b/8$.}
\label{fig:methods_outputgen}
\end{figure}

The step to generate outputs of the traditional spiking neurons can be thought as a mapping between membrane potentials and binary values (see Fig.~\ref{fig:methods_outputgen}(a)). Membrane potentials below threshold (denoted by $\gamma$) are translated to zeros and the values above threshold are translated to ones. Depending on threshold and the maximum value of the membrane potentials (denoted by $b$), there are infinitely many ways to come up with a mapping. Existing works on training SNNs generally choose threshold arbitrarily. However, we found that using a constant threshold does not work well across different learning problems. This is because the maximum range of the membrane potentials changes for different datasets. Hence, the same threshold does not guarantee similar mapping or similar learning performance. We simplify the threshold selection step by setting $\gamma$ as a linear function of $b$ and estimating value of $b$ during training. Specifically, we use an exponential moving average (EMA) to keep track of the membrane potential statistic as in~\citep{jacob2018quantization}. We observe the membrane potentials of spiking neurons on each layer during every batch computation and aggregate their maximum value into the EMA. 

\begin{table*}[!t]
\caption{Column 4-5 shows speech recognition performance on TIMIT and LibriSpeech 100h dataset obtained from the GRUs, LSTMs, traditional SNNs, and SNNs with improved inherent recurrence dynamics (SNNs with ipv dyn). Column 6-7 shows the percentage of zeros in outputs from all architectures while Column 8-9 shows the average number of multiplication operations per inference (avg ops/inf) normalized to that of the LSTMs.}
\label{tab:table1}
\begin{center}
\begin{tabular}{|c|c|c|c|c|c|c|c|c|} 
\hline
\multirow{3}{*}{\shortstack{Architecture}} &
 \multirow{3}{*}{\shortstack{Output\\precision}} &
 \multirow{3}{*}{\shortstack{Thre-\\shold}} & \multicolumn{2}{c|}{Prediction accuracy (\%)} & \multicolumn{2}{c|}{\# of zero outputs (\%)} & \multicolumn{2}{c|}{Avg ops/inf (normalized)} \\
\cline{4-9}
& & & \multirow{2}{*}{~TIMIT~} & LibriSpeech & \multirow{2}{*}{~TIMIT~} & LibriSpeech & \multirow{2}{*}{~TIMIT~} & LibriSpeech \\
& & & & 100h &  & 100h & & 100h \\
\hline
LSTMs & Full & - & $82.68$ & $89.95$ & ${<}0.01$ & ${<}0.01$ & $1.00$ & $1.00$ \\
\hline
GRUs & Full & - & $82.26$ & $89.77$ & ${<}0.01$ & ${<}0.01$ & $0.81$ & $0.83$ \\
\hline
Traditional SNNs & $6$-bit & $b/16$ & $70.66$ & $78.39$ & $59.90$ & $59.96$ & $0.13$ & $0.16$ \\
\hline
SNNs with ipv dyn v1 & $6$-bit & $b/16$ & $79.64$ & $87.36$ & $70.89$ & $67.80$ & $0.15$ & $0.19$ \\
\hline
SNNs with ipv dyn v2 & $6$-bit & $b/16$ & $81.28$ & $88.25$ & $84.29$ & $86.33$ & $0.08$ & $0.08$ \\
\hline
\end{tabular}
\end{center}
\end{table*}

Another problem that can be handled during training is the gradient mismatch arisen from the use of surrogate function to overcome spiking neurons' non-differentiability. Gradient mismatch has been previously shown to be prominent when output precision is low~\citep{lin2016overcoming}.
To make spiking neurons produce multi-bit outputs and to alleviate gradient mismatch issue, we can come up with a multi-level mapping, where membrane potentials are mapped multiple values (as opposed to zeros and ones shown in Fig.~\ref{fig:methods_outputgen}(a)).
We can then apply the proposed technique for threshold selection. As illustrated in Fig.~\ref{fig:methods_outputgen}(b-c), values below the threshold are mapped to zeros. Only sufficiently large membrane potentials lead to non-zero outputs. In order to express the mapping mathematically, we simplify the output computation by imposing two constraints.
First, we limit thew threshold ($\gamma$) to be a multiple of $b/2^{N_{bits}}$ where $N_{bits}$ is the number of bits for output representation.
For example, threshold is set to $2b/4$ in Fig.~\ref{fig:methods_outputgen}(b) where outputs are 2-bit values. Threshold is set to $3b/8$ in Fig.~\ref{fig:methods_outputgen}(c) where outputs are 3-bit values. This constraint helps reducing the space for parameter search during training.
Second, we assume a uniform mapping between membrane potentials and outputs. As illustrated in Fig.~\ref{fig:methods_outputgen}(b), membrane potentials between $[0,b)$ are mapped to $\{0,1,2,3\}$ while the values below $\gamma=2b/4$ are mapped to zeros. 
With the two constraints, the activation function of the spiking neuron can be written as follows:
\begin{align}
    \mathbb{Y}[n] &= 
        \begin{cases}
        \lfloor {\frac{\mathbb{V}[n]}{b}\cdot2^{N_{bits}}} \rfloor & \text{if}~\mathbb{V}[n] > \gamma \\
        0 & \text{if}~\mathbb{V}[n] \leqslant \gamma
        \end{cases}
\end{align}

\section{Experimental Results} \label{sec:experiment}

We demonstrated the effectiveness of the proposed SNNs with improved inherent recurrence dynamics on two speech recognition problems, namely phoneme recognition on TIMIT dataset~\citep{garofolo1993darpa} and word recognition on 100 hours samples of LibriSpeech dataset (LibriSpeech 100h)~\citep{panayotov2015librispeech}. For convenience, we call networks of LIF neurons before the proposed modifications as traditional SNNs throughout this work. We extended PyTorch-Kaldi framework~\citep{ravanelli2019pytorch} to simulate the behavior of SNNs and performed the training.
 
\subsection{Experimental Setup}
Because we target edge devices where the inputs arrive and are processed sequentially, unidirectional LSTMs and GRUs were used as baseline architectures. All architectures used in this section were set up similarly unless otherwise stated. LSTMs, GRUs, and spiking neurons were stacked into different two-layer networks where each layer consisted of $550$ non-spiking or spiking units. Final outputs from those stacked units were fed to a fully connected layer to produce a probability of the most likely phonemes or words for speech recognition purpose. Inputs to neural networks were generated according to Kaldi recipe~\citep{povey2011kaldi}, a popular toolkit for speech recognition. Raw audios were transformed into acoustic features by feature space maximum likelihood linear regression (commonly known as fMLLR) that provides a speaker adaptation. The features were computed using windows of $25$ ms with an overlap of $10$ ms. Before training, we initialized all weight matrices with Glorot's scheme~\citep{glorot2010understanding} except recurrent weight matrices that had orthogonal initialization. We used Adam as an optimizer and recurrent dropout as a regularization technique~\citep{semeniuta2016recurrent}. Dropout rate of $0.1$ was found to give the best performance on all architectures. Batch normalization was applied to control the distribution of weight-sums after multiplying synaptic weights with inputs. Batch size of $64$ was used throughout the training and all architectures were trained for $24$ epochs. The error on the development set was monitored every epoch and the learning rate was halved after $3$ epochs of improvement less than $0.1$. The initial learning rate was set to $1\times10^{-3}$ for all architectures.

\subsection{Speech Recognition Accuracy}

Columns $4$-$5$ of Table~\ref{tab:table1} report the speech recognition performance obtained from LSTMs, GRUs, and SNNs on the TIMIT and LibriSpeech 100h dataset. To account for any variability during training, prediction accuracy was the average of the training results obtained from $5$ experiments with different initial random seeds. We did not constrain the range of outputs for LSTMs and GRUs. In other words, we kept outputs in a floating-point representation during the training and inference. Hence, we denote precision of those architectures at inference as full precision. In case of the SNNs, we limited the outputs to $6$ bits and the threshold to $b/16$ during the training and inference. We arbitrarily chose this setting of SNNs for comparison with baseline architectures; the impact of the output ranges is later discussed in detail. Prediction accuracies from the baseline architectures were lower than previously reported in the literature because we considered only unidirectional models. LSTMs outperformed GRUs by a small margin.
The SNNs with improved dynamics v1 boosted the prediction accuracy of the traditional SNNs by $8.98\%$ and $9.84\%$ on TIMIT and LibriSpeech 100h dataset. This improvement highlighted the benefit of learning without a vanishing gradient and gradient mismatch problem.
The SNNs with improved dynamics v2 further increased the prediction accuracy on top v1 by $1.64\%$ and $5.57\%$ on TIMIT and LibriSpeech 100h dataset.
Determining the forget and candidate signal based on the previous synaptic current thus helped to maximize the recognition accuracy.
With all the proposed architecture and training scheme, the gap in the recognition accuracy between SNNs and the baseline LSTMs essentially became $1.10\%$ and $0.36\%$ on TIMIT and LibriSpeech 100h dataset, with $2\times$ improvement in the number of parameters.

\subsection{Computational Saving}

Columns $6$-$7$ of Table~\ref{tab:table1} report the percentage of zeros in the outputs from all architectures. Only a few outputs from the LSTMs and GRUs were zeros because there was no constraint on how the outputs were generated. SNNs, on the other hand, had many zero outputs because of thresholds that filtered out many small inputs. To support our claim that those zero outputs potentially lead to substantial computation saving on event-driven hardware, we measured the average number of multiplication operations per inference from all architecture and normalized it to the measurement from LSTMs as illustrated in Column $8$-$9$. The operation of the GRUs was cheaper than the operation of LSTMs as GRU computation involved only $3/4$ vector-matrix multiplications of LSTM computation. The simpler dynamics of the traditional SNN and its high output sparsity led to higher computation saving ($5.20$-$6.32\times$ more than the GRUs); however, using the traditional SNN came at a cost of lower recognition accuracy ($11.60$-$15.48\%$ worse than the GRUs).
SNNs with improved dynamics v1 alleviated this shortcoming by reducing the gap in the recognition accuracy between SNNs and GRUs while it maintained the benefit of the sparse computation. Sparse outputs of the proposed SNN led to $4.37$-$5.40\times$ fewer number of multiplications than GRUs.
Note, the traditional SNNs and the SNNs with improved dynamics v1 updated synaptic currents solely based on the inputs. We found that the average number of operations per inference further reduced if spiking neurons updated their synaptic currents as introduced in the SNN with improved dynamics v2. The SNN with improved dynamics v2 had the highest output sparsity (i.e. have highest percentage zeros in the outputs) among all the architectures.
Hence, the SNN with improved dynamics v2 reduced the number of multiplication operations by $10.13\times$ and $11.14\times$ compared to the GRUs for speech recognition task on TIMIT and LibriSpeech 100h datasets, respectively.

\subsection{Effect of the Output Precision on Speech Recognition Accuracy}

\begin{figure}[!t]
\begin{center}
\includegraphics[width=\linewidth]{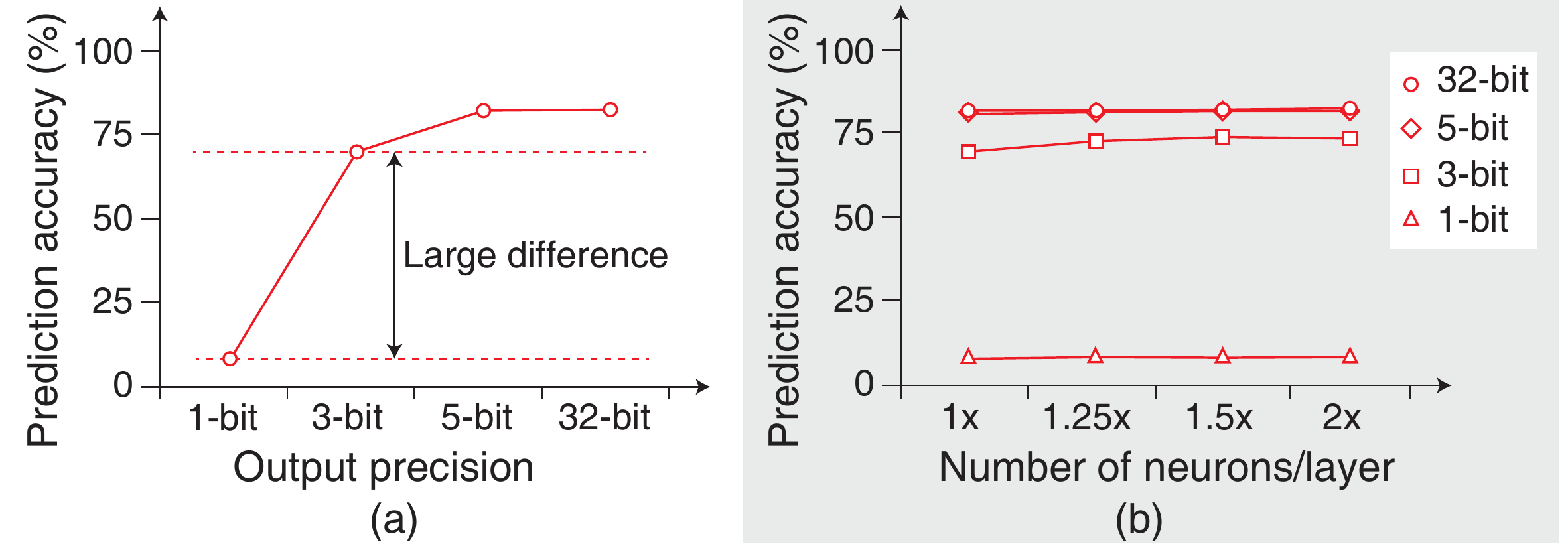}
\end{center}
\caption{Speech recognition performance of the the SNN with improved dynamics v2 on TIMIT dataset: (a) prediction accuracy with various output precisions and (b) prediction accuracy with various output precisions and network sizes.}
\label{fig:results_acc_comp}
\end{figure}

Fig.~\ref{fig:results_acc_comp}(a) compares the speech recognition performance obtained from the SNN with improved dynamics v2 with various output precisions. 
The threshold was assumed to be $b/2^{N_{bits}}$ in each case to minimize the effect of the threshold on the output generation step and to follow our proposed simplification that limits threshold to be a multiple of $b/2^{N_{bits}}$ where $N_{bits}$ is the number of bits for output representation.
For instance, the threshold was set to $b/2$ for 1-bit outputs and was set to $b/8$ for 3-bit outputs.
Reducing the number of output precision from 3-bit to 1-bit led to a large drop in the accuracy of the SNN from $70\%$ to $8.8\%$. Once we increased the output precision to 4 bits or more, we achieved the recognition accuracy close to training accuracy without any constraint on the outputs.
One may argue that the sharp accuracy drop occurs because of the limited capacity of the SNN with binary outputs that is not large enough for a given task. The capacity of the SNN with binary outputs may be improved by increasing more neurons per layer and allow the network to learn successfully. Hence, we conducted another set of experiments by varying the size of the spiking neurons in each layer to increase the network capacity.
As illustrated in Fig.~\ref{fig:results_acc_comp}(b), the large accuracy drop still existed between the SNN with binary and $3$-bit outputs. Even if the number of spiking neurons is increased by $2\times$ per layer, the SNN with binary outputs still performed poorly compared to the SNN with 3-bit outputs. Similar results were also observed from a speech recognition task on Librispeech 100h dataset. We used these experimental results to argue that sharp accuracy drop for binary output stems from inefficiency in the training method rather than the limited capacity of SNN. The experimental results also showed that increasing output precision is one way to improve the recognition accuracy of the SNN on sequential learning.

\section{Conclusion} \label{sec:conclusion}

SNNs have been considered as a potential solution for the low-power machine intelligence due to their event-driven nature of computation and the inherent recurrence that helps retain information over time. However, practical applications of SNNs have not been well demonstrated due to an improper task selection and the vanishing gradient problem. In this work, we proposed SNNs with improved inherent recurrence dynamics that are able to effectively learn long sequences. The benefit of the proposed architectures is $2\times$ reduction in number of the trainable parameters compared to the LSTMs. Our training scheme to train the proposed architectures allows SNNs to produce multiple-bit outputs (as opposed to simple binary spikes) and help with gradient mismatch issue that occurs due to the use of surrogate function to overcome spiking neurons' non-differentiability.
We showed that SNNs with improved inherent recurrence dynamics reduce the gap in speech recognition performance from LSTMs and GRUs to $1.10\%$ and $0.36\%$ on TIMIT and LibriSpeech 100h dataset.
We also demonstrated that improved SNNs lead to $10.13$-$11.14\times$ savings in multiplication operations over standard GRUs on TIMIT and LibriSpeech 100h speech recognition problem. This work serves as an example of how the inherent recurrence of SNNs can be used to effectively learn long temporal sequences for applications on edge computing platforms.

\bibliography{refs}

\end{document}